\documentclass[letterpaper, 10 pt, journal, twoside]{ieeetran}
\usepackage{amsmath, amssymb}
\usepackage{dsfont}
\usepackage{tablefootnote}

\usepackage{cite}
\usepackage{hyperref}
\usepackage{url}
\usepackage{float}
\usepackage{graphicx}
\usepackage{tikz}
\usepackage{subcaption}
\usepackage{pifont}
\usepackage{siunitx}
\usepackage{amssymb}

\usepackage{pifont}
\usepackage{makecell}
\usepackage{wrapfig}
\usepackage{caption}  
\usepackage{arydshln}

\usepackage{booktabs}
\usepackage{amsmath}
\usepackage{multirow}
\usepackage{float}
\usepackage{graphicx}
\usepackage{tikz}
\usepackage{subcaption}
\usepackage{pifont}
\usepackage{siunitx}
\usepackage{verbatim}
\usepackage{makecell}
\def\BibTeX{{\rm B\kern-.05em{\sc i\kern-.025em b}\kern-.08em
    T\kern-.1667em\lower.7ex\hbox{E}\kern-.125emX}}
\usepackage{balance}

\begin{document}
\title{Robust Scene Transfer for PointGoal Navigation via Privileged-Sensor–Guided Contrastive Learning}
\author{Amirhossein Zhalehmehrabi, Tiziano Tezze, Alberto Castellini, and Alessandro Farinelli
\thanks{Manuscript received: March 16, 2026; Revised June 23, 2026; Accepted August 2, 2026. This paper was recommended for publication by Editor P. Vasseur upon evaluation of the Associate Editor and Reviewers' comments. This work was supported by PNRR iNEST project ECS00000043}
\thanks{The authors are with the Department of Computer Science, University of Verona, Italy. Email: firstname.lastname@univr.it}
\thanks{Digital Object Identifier (DOI): see top of this page.}
}

\markboth{IEEE Robotics and Automation Letters. Preprint Version. Accepted August, 2026}
{Zhalehmehrabi \MakeLowercase{\textit{et al.}}: Robust Scene Transfer via Privileged-Sensor Contrastive Learning} 

\maketitle

\begin{abstract}
We propose a sensor-guided adaptive contrastive learning framework for visual representation learning in PointGoal navigation. During training, privileged LiDAR sensing guides the contrastive objective through a geometry-aware similarity metric and adaptive temperature scaling, encouraging visual embeddings to capture navigation-relevant structure rather than scene-specific appearance. The resulting encoder is pretrained independently, frozen, and used as the perceptual backbone for reinforcement learning, decoupling representation learning from policy optimization. We further introduce a cross-stage domain mismatch between representation pretraining and policy learning to suppress environment-specific shortcuts and promote reliance on task-relevant features.

Extensive experiments in high-fidelity simulation demonstrate that our approach significantly improves policy-level scene transfer across diverse indoor and outdoor environments. At deployment, the agent relies only on monocular RGB observations together with standard task-related inputs such as goal position and proprioceptive signals, without access to LiDAR or other privileged sensors. Our method outperforms large pretrained vision models, navigation-specific representation learning approaches, domain randomization methods, and standard contrastive baselines under severe appearance and semantic shifts. We also release a multimodal dataset to support future research on privileged-guided visual representation learning for navigation. The code is available at: \href{https://github.com/zhalehmehrabi/privileged-sensor-contrastive-nav}{https://github.com/zhalehmehrabi/privileged-sensor-contrastive-nav}
\end{abstract}

\section{Introduction}

Vision-based mobile navigation promises scalable autonomy by enabling robots to operate using only onboard cameras, avoiding reliance on additional sensing modalities such as depth or LiDAR. Despite this appeal, vision-based navigation policies often fail when deployed in visually different environments, even when task dynamics and robot embodiment remain unchanged \cite{kaufmann2023champion}. This lack of \emph{scene transfer}—the ability to reliably perform the same task across visually distinct environments—remains a major barrier to real-world deployment of learning-based navigation systems.

\begin{figure}[]
    \centering
    \includegraphics[width=0.5\textwidth]{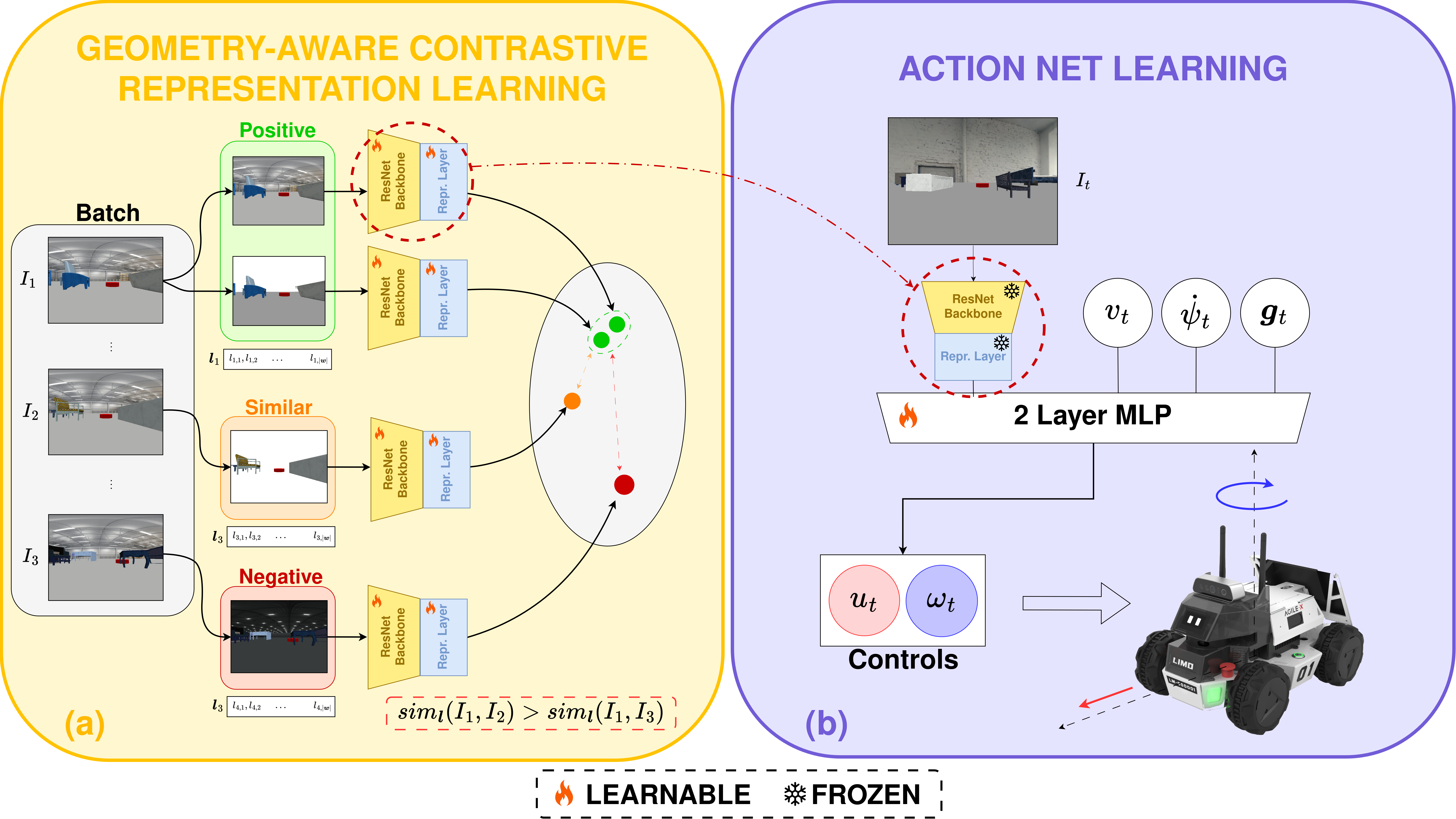}
    \caption{Overall pipeline. (a) A ResNet backbone is first trained with geometry-aware contrastive learning, (b) then frozen and used alongside robot state inputs to train an MLP policy that outputs navigation controls.}
    \label{fig:pipeline}
\end{figure}

At the core of this challenge lies representation learning. To generalize across environments, a navigation policy must rely on visual representations that suppress scene-specific appearance while preserving task-relevant structure such as goal geometry, free space, and obstacle layout. While prior work has demonstrated strong performance within fixed environments \cite{fifty2021efficiently, shridhar2023perceiver, rahmatizadeh2018vision}, learning representations that generalize across environments without retraining remains challenging, particularly under large visual distribution shifts.

In contrast to robotic manipulation, where scene transfer has been extensively studied \cite{zeng2022robotic}, vision-based navigation methods are typically evaluated in visually similar environments or rely on explicit geometric representations, such as mapping, SLAM, or depth sensing \cite{wurman2022outracing, du2021curious, Bonanni_AAMAS_2025, Taioli_2025_ICCV}.

In another line of research, contrastive representation learning provides a principled mechanism for enforcing invariance by aligning semantically similar observations \cite{chen2020simple}. However, standard contrastive methods rely on heuristic augmentations and do not explicitly incorporate task structure, often resulting in representations that remain entangled with scene appearance under large cross-environment shifts \cite{wang2021understanding}.

Recent work \cite{xing2024contrastive} incorporates privileged supervision into contrastive learning for agile flight. However, this approach is trajectory-centric and demonstrated only for executing predefined motion patterns, limiting its applicability to general closed-loop navigation. Moreover, reliance on imitation learning reduces robustness in long-horizon tasks due to limited recovery outside the training distribution \cite{luo2025precise}.

These limitations highlight a key gap: policy-level scene transfer requires representations that support scene-agnostic decision-making. More broadly, this work explores a research direction in which privileged sensing modalities guide the representation learning of deployable sensors. Rather than relying on these sensors at inference time, we use them during training to shape the feature extractor of the target sensor. One promising direction is to leverage sensing modalities that provide direct access to geometric structure, which is inherently less sensitive to appearance variations across environments. In this work, we instantiate this idea by using LiDAR observations to guide the learning of an RGB encoder. In this work, we use privileged geometric sensing available during training to guide representation learning, without requiring such sensors at deployment. Specifically, we incorporate geometric observations into the contrastive objective as an additional supervisory signal, encouraging the learned representation to encode the underlying geometric structure of the scene. This geometry-guided regularization steers the representation away from appearance-specific features and toward features that remain stable across environments, resulting in more robust and transferable policies.

Specifically, we propose a \emph{sensor-guided adaptive contrastive learning} framework in which privileged LiDAR observations modulate the contrastive objective through geometry-aware adaptive temperature scaling. The resulting visual encoder is pretrained independently, frozen, and used as the perceptual backbone for reinforcement learning (RL). At deployment, the navigation policy operates using only deployable sensing modalities, including monocular RGB observations and standard task inputs such as goal-relative pose and proprioceptive measurements, without access to the privileged sensing used during representation pretraining.

We summarize our contributions as follows:
\begin{itemize}
    \item \textbf{Sensor-guided adaptive contrastive learning.}
    We propose a framework in which privileged sensors guide the representation learning of a target sensor by modulating the contrastive objective. We instantiate this paradigm using LiDAR observations to guide an RGB encoder.

    \item \textbf{Policy-level scene transfer with decoupled pretraining and RL.}
    We demonstrate that freezing a sensor-guided pretrained visual-based encoder and training navigation policies via RL yields robust closed-loop generalization across visually distinct environments.

    \item \textbf{Comprehensive evaluation and dataset release.}
    We demonstrate strong cross-scene generalization on PointGoal navigation and release a multimodal dataset to support future research on privileged-guided visual representation learning.
\end{itemize}
\section{Related Works}

\subsection{End-to-end Policy Learning}

End-to-end sensorimotor learning jointly optimizes perception and control directly from raw sensory observations, bypassing explicit geometric modeling \cite{levine2016end}. In PointGoal navigation, map-based approaches integrate spatial representations and planning, while mapless approaches learn policies directly from sensory inputs. However, mapless methods typically rely on rich sensing and task rewards as primary supervision, which does not explicitly enforce scene-invariant representations and often leads to visual overfitting.

Our work explicitly targets scene transfer in vision-based navigation by decoupling representation learning from policy optimization—learning task-relevant visual features first, then training a policy on the frozen representation. Unlike prior geometry-driven approaches \cite{zhalehmehrabi2025depthconstrained, xiao2022motion}, we learn transferable features through contrastive pretraining.

\subsection{Visual Pre-training for Robotics}

Visual pre-training has emerged as an effective strategy for improving generalization and sample efficiency in robotic learning by enabling reusable representations \cite{radosavovic2023real}. Large-scale pretrained models such as CLIP \cite{radford2021learning} and Masked Autoencoders (MAE) \cite{he2022masked} have demonstrated strong cross-task transfer in robotic perception and control \cite{shridhar2022cliport}.

Self-supervised and contrastive learning methods provide a complementary approach by shaping representations according to task dynamics rather than raw appearance. Contrastive pretraining, popularized by methods such as SimCLR \cite{chen2020simple}, as well as subsequent extensions in robotics \cite{messikommer2024contrastive}, and contrastive RL approaches \cite{qiu2022contrastive}, improve robustness by learning invariant, task-relevant features. Representative paradigms include trajectory-based contrastive learning, which pulls future states closer \cite{eysenbach2022contrastive}, and augmentation-based methods that enforce consistency across transformed observations \cite{du2021curious}.

\cite{xing2024contrastive} proposed an adaptive contrastive learning framework guided by privileged signals in a controlled flight setting. However, this approach relies on predefined trajectories and teacher supervision, limiting its applicability to autonomous navigation. In contrast, our method uses auxiliary sensory information available only during training to adaptively modulate the contrastive objective, without requiring privileged environmental states or teacher policies.

\subsection{Privileged Visual Representation Learning}

Privileged supervision leverages sensors or state information available during training but not at deployment to facilitate representation learning and policy optimization. In robotics, this paradigm has been widely used in teacher–student or “learning-by-cheating” frameworks, where controllers trained with full-state or geometric inputs are distilled into vision-based policies \cite{levine2016end}.

More recently, privileged information has been integrated into self-supervised and contrastive learning by guiding sampling strategies or modulating contrastive objectives \cite{ xing2024contrastive}. These approaches typically rely on trajectory structure, expert supervision, or explicit state regression, which restricts their applicability to general closed-loop navigation.

In contrast, our approach uses privileged geometric sensing solely during training to adaptively scale a self-supervised contrastive objective, without teacher policies or fixed trajectories.
\section{Methodology}
We propose a vision-based robot navigation framework that separates visual representation learning from policy optimization. The visual encoder is pretrained with a privileged-sensor--guided contrastive objective using replay correspondences and geometric observations available only during representation learning. Once pretraining is complete, the encoder is frozen and used as a feature extractor for a lightweight policy trained via model-free RL. No privileged observations, replay trajectories, or cross-scene correspondences are used during policy training or evaluation.

\subsection{Task Formulation}
\label{sec:setup}
We study the PointGoal Navigation task, where a robot must reach a target relative to its start, avoiding obstacles. At each timestep, it receives partial observations and outputs continuous linear and angular velocities. Episodes end on success or collision. During deployment, the robot observes the goal in its local frame, velocities, and an RGB image, without access to obstacle geometry or absolute position. The aim is to learn a navigation policy that generalizes across visually diverse environments using only these sensors.

\subsection{Geometry-Aware Contrastive Representation Learning}
\label{sec:contrastive}

The objective of the representation learning stage is to train a vision encoder that produces embeddings aligned with navigation-relevant geometry rather than scene appearance. We build upon standard contrastive learning, but incorporate privileged geometric information available during training to modulate the loss in a task-aware manner.

Training batches are sampled uniformly. Positives are augmented views of scene-invariant counterparts, while all other samples in the batch are treated as negatives. Instead of altering this pair assignment, we adjust the contribution of each pair to the contrastive objective through a geometry-adaptive temperature mechanism.

\subsubsection{Contrastive Learning Objective}
Let $\phi_{\boldsymbol{\theta}}$ denote a vision encoder parameterized by $\boldsymbol{\theta}$ that maps an RGB image $I_i$ to a latent embedding $\mathbf{z_i} = \phi_{\boldsymbol{\theta}}(I_i) \in \mathbb{R}^d$. Given a random batch of $N$ images, we train $\phi_{\boldsymbol{\theta}}$ to minimize the \emph{InfoNCE objective} \cite{vandernoord2018infonce} for each positive pair in the batch:
\begin{align}
    \mathcal{L}_{i,j} = -\log \frac{\exp\Big(\textrm{sim}(\mathbf{z_i}, \mathbf{z_j}) / \tau\Big)}{\sum_{k=1}^{N} \mathds{1}_{[k \neq i]} \exp\Big(\textrm{sim}(\mathbf{z_i}, \mathbf{z_k}) / \tau\Big)} \label{eq:snn}
\end{align}
where $\mathrm{sim}(\cdot,\cdot)$ denotes cosine similarity and $\tau$ is a temperature parameter. In contrast to standard formulations, $\tau$ is not fixed but adaptively modulated based on geometric similarity, as described in Section~\ref{sec:adaptive_tau}.

\subsubsection{Task-Aware Similarity Metrics}
\label{sec:sim}
Although pair assignment follows standard contrastive practice, uniform treatment of negatives is suboptimal for navigation. Visually distinct observations may correspond to nearly identical local geometric configurations, and forcing strong repulsion between such embeddings conflicts with the objective of learning geometry-aligned representations.

To address this, we compute a geometry-based similarity using LiDAR readings $\boldsymbol{l}$ available during training. Given two observations $I_1$ and $I_2$ with corresponding vectors $\boldsymbol{l}_1$ and $\boldsymbol{l}_2$, we define their \emph{normalized geometric similarity} as
\begin{align}
sim_{\boldsymbol{l}}(I_1, I_2) = 
1 - \frac{\sqrt{\sum_{i=1}^{\ell} w_i (\boldsymbol{l}_{1,i} - \boldsymbol{l}_{2,i})^2}}
{\sqrt{\sum_{j=1}^{\ell} w_j}}
\label{eq:dz}
\end{align}
where $\ell=|\boldsymbol{l}|$ is the number of LiDAR beams, and $w_i \geq 0$ is the weight assigned to the $i$-th beam; their selection is detailed in Section~\ref{sec:imple-vis-enc}.

The subtracted term corresponds to a normalized weighted Euclidean distance in LiDAR space, bounded in $[0,1]$, such that higher values of $sim_{\boldsymbol{l}}$ indicate greater similarity in the local spatial structure surrounding the robot.

\subsubsection{Geometry-Adaptive Temperature}
\label{sec:adaptive_tau}
Contrastive learning objectives are known to be sensitive to the temperature parameter \cite{wang2021contrastivelossbehavior,xing2024contrastive,zhuang2024softcl}, which controls the strength of attraction and repulsion in embedding space. We therefore modulate the temperature on a per-pair basis according to geometric similarity, allowing repulsive forces to vary continuously with navigation-relevant structure.

Given a contrastive pair $(I_1, I_2)$ with geometric similarity $sim_{\boldsymbol{l}}(I_1,I_2)$, we define the adaptive temperature as
\begin{align}
\tau(I_1,I_2) &=
\begin{cases}
\alpha & \small \text{if $(I_1,I_2)$ is a positive pair}, \\
sim_{\boldsymbol{l}}(I_1,I_2) & \small \text{if $(I_1,I_2)$ is a negative pair}.
\end{cases}
\end{align}
The resulting value is mapped to a bounded interval $[\tau_{\min}, \tau_{\max}]$ to ensure numerical stability:
\begin{align}
\tau'(I_1,I_2) = \tau_{\min} + (\tau_{\max} - \tau_{\min}) \cdot \tau(I_1,I_2)
\label{eq:tau_map}
\end{align}

This adaptive formulation ensures that geometrically similar observations, even when treated as negatives, apply weaker repulsive forces, while geometrically dissimilar observations are pushed apart more strongly. As a result, the learned latent space reflects navigation-relevant structure rather than purely visual appearance. In our framework, privileged LiDAR observations are used exclusively during simulation-based representation pretraining and are never available at deployment. Since privileged sensing operates entirely within the simulator, geometric measurements are noise-controlled by construction, and we assume perfect sensing during training. The effect of degraded privileged sensing is therefore outside the scope of this work.

\begin{figure}
    \centering
    \includegraphics[width=0.5\textwidth]{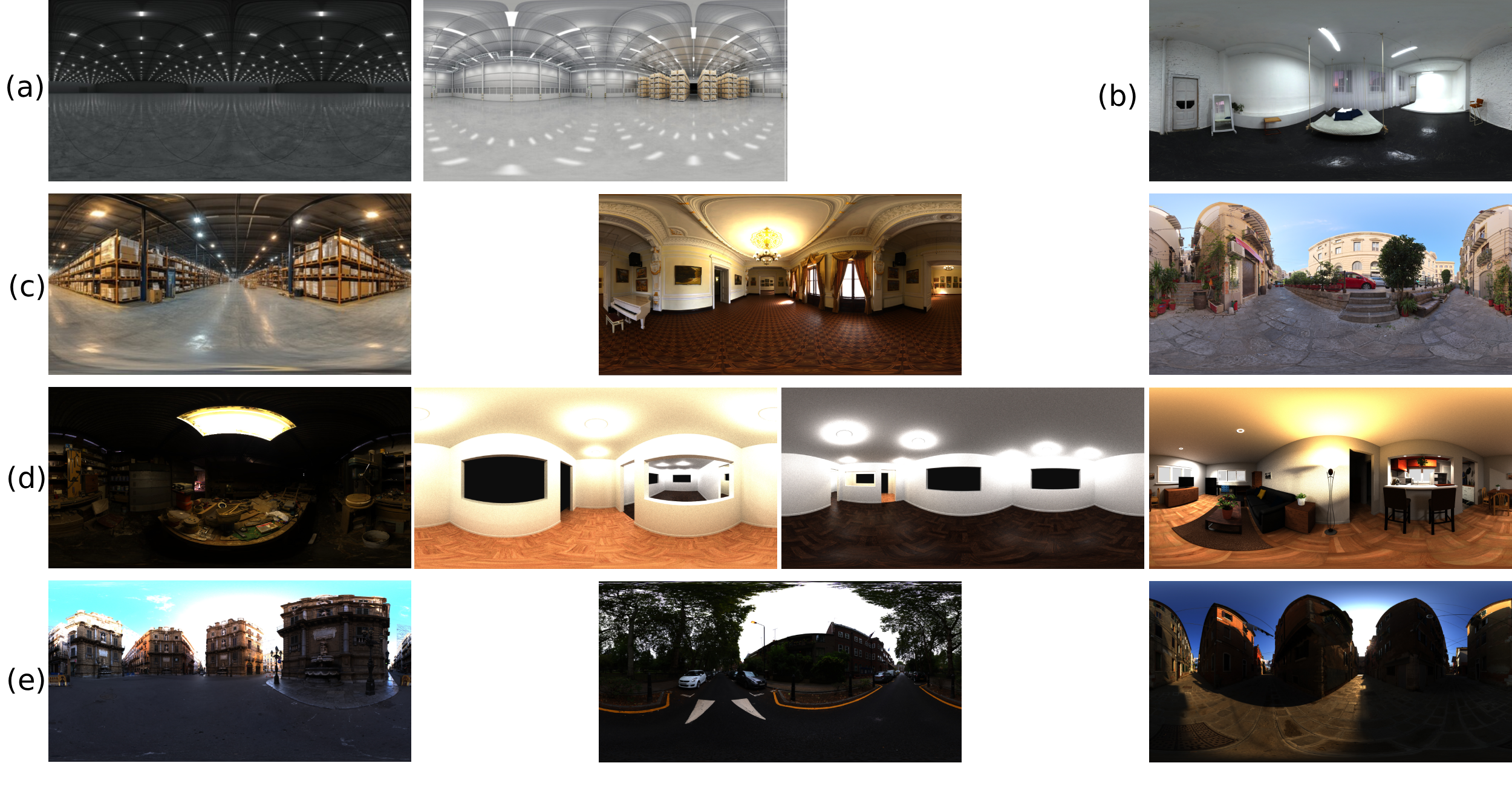}
    \caption{Overview of datasets and evaluation environments. Visual appearance differences are shown; obstacle counts and placements are not depicted. (a) Environments for visual encoder pre-training, Warehouse-1\&2 (including variants without background and without floor/background).
(b) RL training environment, Photo-studio.
(c) Environments for representation analysis, Warehouse-3, Ballroom, Palermo.
(d) Indoor test environments, Carpentry-shop, Probe-2\&3, Rs.
(e) Outdoor test environments, Quattro-canti, Urban-street, Venetian-crossroads.}
    \label{fig:data_overview}
\end{figure}


\subsection{Action Net Learning}
\label{sec:action_net_learning}
The navigation policy is composed of a frozen vision encoder and a lightweight action network. At each timestep $t$, the policy receives an observation $o_t$ consisting of the visual embedding $\mathbf{z_t} = \phi_{\boldsymbol{\theta}}(I_t)$ extracted from the RGB image $I_t$, the robot’s linear and angular velocities, $v_t, \dot{\psi}_t$ respectively, and the relative position of the goal in polar coordinates $\mathbf{g}_t$. Formally, the observation is defined as:
\begin{equation}
    o_t = \big\{ \mathbf{z_t}, \, v_t, \, \dot{\psi}_t, \, \mathbf{g}_t \big\}
\end{equation}
The action network maps observation to continuous control commands: linear velocity $u_t$ and angular velocity $\Omega_t$.

The action network is optimized using a model-free RL algorithm. By fixing the visual representation during policy learning, the action network is trained on a stable and scene-invariant input distribution, which helps reduce overfitting to the training environments and isolates the effect of representation learning on policy generalization.

The reward function is inspired by \cite{partsey2022mapping} and is defined as:
\begin{equation}\label{eq:reward}
        \footnotesize R(s_t, a_t, s_{t+1}) =
        \begin{cases}
        r_{\text{suc}} & \text{\footnotesize if success}, \\
        r_{\text{collide}} & \text{\footnotesize if collision}, \\
        -\alpha_1 \Delta_d - \alpha_2 r_{\text{back}} - \alpha_3 r_{\text{angular}} & \text{\footnotesize otherwise},
        \end{cases}
\end{equation} 
where $\Delta_d$ denotes the change in geodesic distance to the goal between consecutive states, $r_{\text{back}} = |\min(u_t, 0)|$ penalizes backward motion, and $r_{\text{angular}} = |\Omega_t|$ penalizes angular velocities. Penalizing angular velocity encourages smoother trajectories and more stable visual observations, which improves robustness when operating with vision-based perception. The reward is kept identical across all environments.

\subsection{Cross-Stage Domain Mismatch (CSDM)}
\label{sec:csdm}
The environments used for representation pretraining (Fig.~\ref{fig:data_overview}~(a)) differ from those used during action net learning (Fig.~\ref{fig:data_overview}~(b)). This cross-stage domain mismatch is intentional and serves to isolate the effect of representation learning on policy generalization. By freezing the pretrained encoder, the visual feature space remains fixed during RL. Training the policy in a distinct environment discourages reliance on scene-specific correlations that may arise during pretraining, as such correlations do not transfer across environments. Consequently, policy optimization is forced to exploit features that remain predictive under domain shift, providing a stronger test of geometry-aligned representation learning.

\section{Dataset and Data Collection}

We collect navigation episodes by replaying trajectories from a privileged \emph{state-based} agent—trained with full geometry access but no visual inputs—across visually distinct environments sharing identical geometry, yielding observations of the same navigation-relevant configurations under varying appearances.

At each timestep, we record RGB images, LiDAR-derived vectors, depth images, and semantic segmentation masks. For pretraining, only the LiDAR-derived vector is used to define task-relevant similarity in the contrastive objective; the remaining modalities are included for completeness.

The dataset spans diverse indoor and outdoor environments, including three warehouse layouts, Photo-studio, ballroom, Palermo-sidewalks, and no-background/floor variants (Table~\ref{tab:dataset}). To increase scene diversity, we generated 20 room configurations for each environment. Each configuration specifies the spatial arrangement, orientation, and object category assigned to each placement location. Every configuration contains 10 obstacle objects. During data collection, object instances are sampled from a pool of 126 assets belonging to 13 categories, allowing different object appearances to occupy the same semantic role across episodes while preserving the underlying room layout. Object placements therefore vary across room configurations, while object instances are randomized within each category. Four environments are used for representation pretraining (Fig.~\ref{fig:data_overview}~(a)), where one warehouse serves as the anchor and others provide scene-invariant counterparts. The Photo-studio environment is reserved for RL training (Fig.~\ref{fig:data_overview}~(b)), and three environments are held out entirely for out-of-distribution validation (Fig.~\ref{fig:data_overview}~(c)). 
All privileged information is used solely during representation learning and is never available to the navigation policy at deployment \footnote{A subset of the GRAN dataset (Geometry-Guided Representation for Autonomous Navigation) is available for download at the following link. The full dataset will be released upon paper acceptance. \scriptsize \textcolor{blue}{mega.nz/file/c6MACbqJ\#CQx3FXBnWfivtEqHyKcxGBWmgg9N1LcvVUmVJDbxkcw}}.

\begin{table}[t]
\centering
\caption{Dataset Specifications.}
\label{tab:dataset}
\small
\resizebox{0.5\linewidth}{!}{
\setlength{\tabcolsep}{4pt}
\begin{tabular}{c c}
\hline
\textbf{Parameter} & \textbf{Value} \\
\hline
\#Environments & 8 \\
\#Images per Env. & 36k \\ 
\#Objects in Env. & 10 \\
\#Room Settings & 20 \\
\#Object Categories & 13 \\
\#Distinct Objects & 126 \\
\#Source Models & 4 \\
\#Trajectories per Model & 100 \\
\end{tabular}
}
\end{table}
\section{Experiments}
We evaluate our approach through representation analysis and downstream navigation under distribution shift.

\subsection{Implementation Details and Model Training}

\subsubsection{Vision Encoder Pre-Training}
\label{sec:imple-vis-enc}
Using the high-fidelity simulator \textit{iGibson} \cite{shen2021igibson}, we collected a dataset for pretraining. We conduct simulation experiments using TurtleBot platforms with a differential-drive configuration.
The visual encoder is trained using the geometry-adaptive contrastive objective described in Sec.~\ref{sec:contrastive}, including the scene-invariant pairing strategy. No additional image augmentations are applied. Architecture details and optimization hyperparameters are reported in Table~\ref{tab:hyperparams}.
For weights $\boldsymbol{w}$ in Eq.~\ref{eq:dz}, beams inside the FOV are assigned unit weight $w_i=1$.
Outside the FOV, weights decay smoothly and symmetrically toward $0.1$ via a sigmoid.

\begin{table}[t]
\centering
\caption{Training hyperparameters for visual pretraining, representation analysis and policy learning.}
\label{tab:hyperparams}
\small
\resizebox{0.8\linewidth}{!}{
\setlength{\tabcolsep}{4pt}
\begin{tabular}{l l l}
\hline
\textbf{Stage} & \textbf{Parameter} & \textbf{Value} \\
\hline
\multirow{7}{*}{Pretrain}
 & RGB Input dim & $224 \times 224$ \\
 & Backbone & ResNet-50 (IN) \\
 & \#Projection layers & 2 \\
 & Projection dim $d$ & 256 \\
 & Batch size & 8192 \\
 & Optimizer & Adam \\
 & LR & $1\mathrm{e}{-4}$ (linear decay) \\
 & $\tau_{\min} / \tau_{\max}$ & $0.1 / 10$ \\
 & $\alpha$ (pos. temp.) &0.1 \\

\hline
\multirow{1}{*}{Representation analysis}
 & k-NN k & 10 \\

\hline
\multirow{6}{*}{RL (SAC)}
 & $r_{suc}$ / $r_{collide}$ & 10 / -10 \\
 & $\alpha_1$ / $\alpha_{2,3}$ & 1 / -0.1 \\
 & Actor / Critic LR & $3\mathrm{e}{-4}$ / $3\mathrm{e}{-4}$ \\
 & Batch size & 1024 \\
 & Replay buffer & $1.5\mathrm{e}{6}$ \\
 & Discount $\gamma$ & 0.99 \\
 & Target update $\tau$ & 0.005 \\
 & Entropy coeff. & automatic \\
\hline
\end{tabular}
}
\end{table}

\subsubsection{Action Network Learning}

The navigation policy consists of the frozen visual encoder and a two-layer MLP with hidden dimension 256. No recurrent components are used to isolate the effect of representation quality. Policy optimization uses Soft Actor-Critic (SAC)~\cite{haarnoja2018soft}. Policy training is conducted exclusively in the Photo-studio environment (Fig.~\ref{fig:data_overview}~(b)), which is not seen during encoder pretraining.

For evaluation, we use seven environments unseen during visual pretraining, representation analysis, and policy learning: four indoor scenes (carpentry-shop, probe-2, probe-3, Rs; Fig.~\ref{fig:data_overview}~(d)) and three outdoor scenes (quattro-canti, urban-street-01, venetian-crossroads; Fig.~\ref{fig:data_overview}~(e)). Each environment is tested under uniform- and textured-floor variants to induce controlled appearance and semantic shifts.

\subsubsection{Baselines}
We compare against three categories of visual representation baselines: (i) frozen pretrained models not exposed to our dataset, including ResNet50~\cite{he2016deep}, CLIP-ResNet/ViT~\cite{radford2021learning}, MAE~\cite{he2022masked}, DINOv2~\cite{oquab2023dinov2}, and CRL~\cite{du2021curious}, a navigation-specific curiosity-driven contrastive representation; (ii) representations trained on our custom dataset, namely AutoEncoder (AE)~\cite{hinton2006reducing} with a ResNet50 backbone, MAE with pretrained weights, and SimCLR~\cite{chen2020simple}, which incorporate domain randomization at the representation-learning stage; and (iii) two SimCLR-based policy-stage domain randomization baselines, DR-PR (trained on the same environments used for pretraining) and DR-CSDM (trained on held-out environments, inducing a cross-stage domain mismatch). Direct comparison with \cite{xing2024contrastive} was not possible, as their approach assumes predefined trajectories for both training and evaluation.

\subsection{Representation Analysis}
We evaluate representations on both training and hold-out scenes illustrated in Fig.\ref{fig:data_overview}~(a) and Fig.\ref{fig:data_overview}~(c). Hold-out scenes are used to validate cross-scene generalization.
\subsubsection{Embedding–Action Distance Correlation}

To quantify whether the learned representation preserves task-relevant geometry, we measure the correlation between distances in the embedding space and corresponding differences in control commands. The idea behind this analysis is that a good encoder should generate distant embeddings for pairs of images requiring different actions, and close embeddings for pairs of images requiring similar actions. This evaluation is inspired by representational similarity analysis \cite{kornblith2019similarity} and prior studies examining geometric structure in learned embeddings \cite{frosst2019analyzing}. Unlike classification-based probes, this metric directly evaluates alignment between latent structure and downstream control signals, which is critical for sensorimotor transfer.

As shown in Fig.~\ref{fig:cor_knn}~(a), our method achieves substantially higher control alignment than all baselines in both training and hold-out scenes. Importantly, the correlation remains stable under scene shift, whereas general-purpose visual encoders exhibit significant degradation. These results indicate that the proposed objective induces a representation whose geometry reflects control-relevant structure rather than purely visual similarity, enabling transfer beyond the visual distribution observed during training.

\subsubsection{k-Nearest Neighbor Action Prediction}

We further assess local control consistency using a non-parametric $k$-nearest neighbor (kNN) regression probe. For a query embedding, control commands are predicted as the average of its $k$ nearest neighbors retrieved via cosine similarity, and performance is measured using mean absolute error. This frozen kNN protocol is commonly used to evaluate representation quality independent of task-specific fine-tuning \cite{caron2021emerging, sermanet2018time}.

\begin{figure}[]
    \centering
\includegraphics[width=0.4\textwidth]{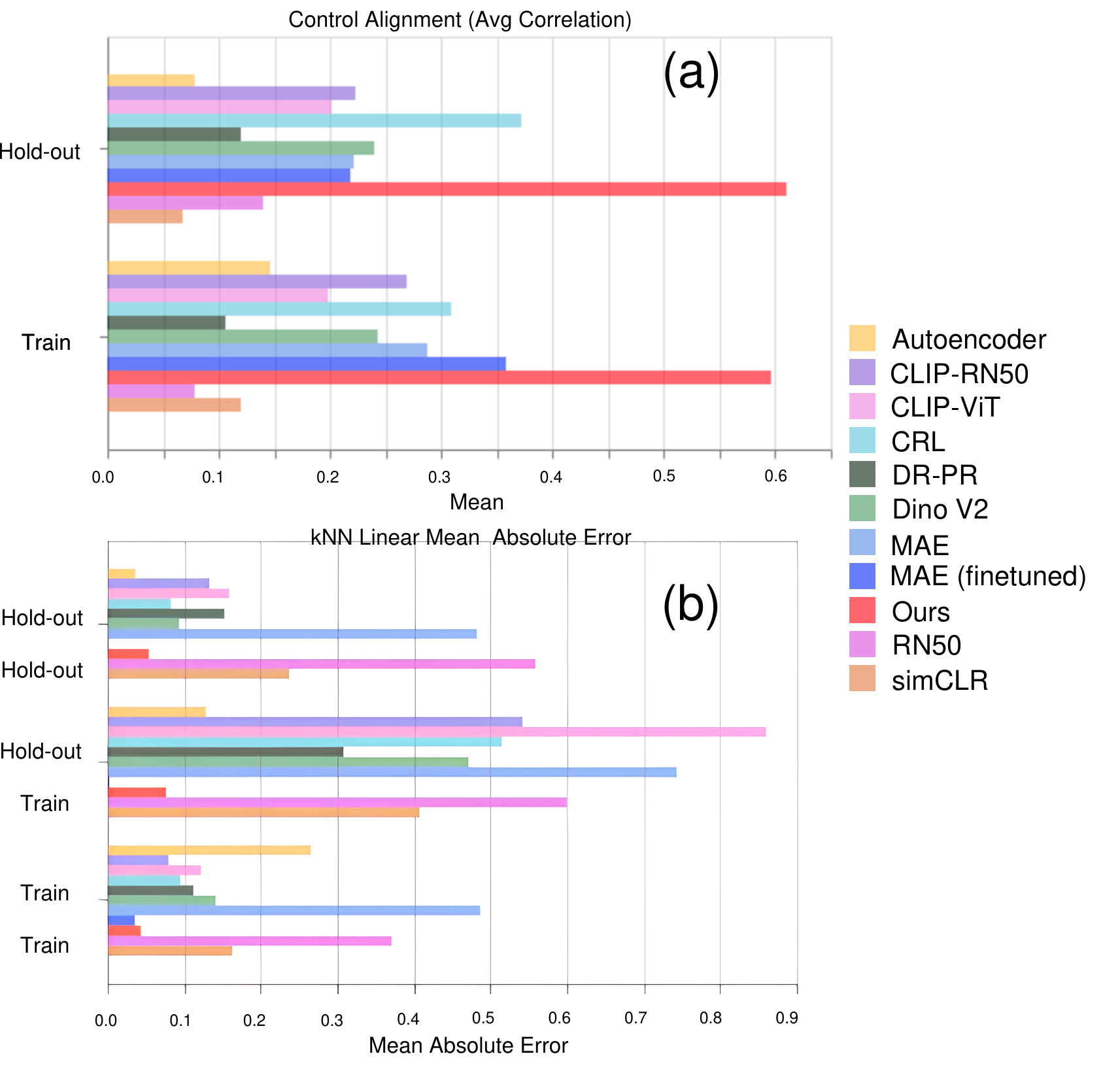}
\caption{Control alignment and local consistency across scenes. (a) Correlation between embedding distances and control differences (higher is better). (b) k-NN action prediction mean absolute error across scene pairs (lower is better), evaluating local control consistency and transferability.}
    \label{fig:cor_knn}
\end{figure}

Results in Fig.~\ref{fig:cor_knn}~(b) show that our representation achieves consistently low kNN mean absolute error across scene combinations, with limited degradation under scene shift. The MAE-finetuned encoder may reach impressive kNN error rates, but these gains are deceptive. The representation lacks a robust embedding–action correlation (Fig. \ref{fig:cor_knn}~(a)) and, as evidenced in Sec. \ref{sec:navigation_performance}, ultimately fails to generalize.

This discrepancy suggests that the finetuned encoder may overfit to dataset-specific local statistics, yielding favorable nearest-neighbor regression performance without preserving globally consistent control-aware geometry. In contrast, our method maintains both strong global alignment and stable local structure, supporting robust transfer.

\subsubsection{U-Map visualization}

\begin{figure}[]
    \centering
\includegraphics[width=0.25\textwidth]{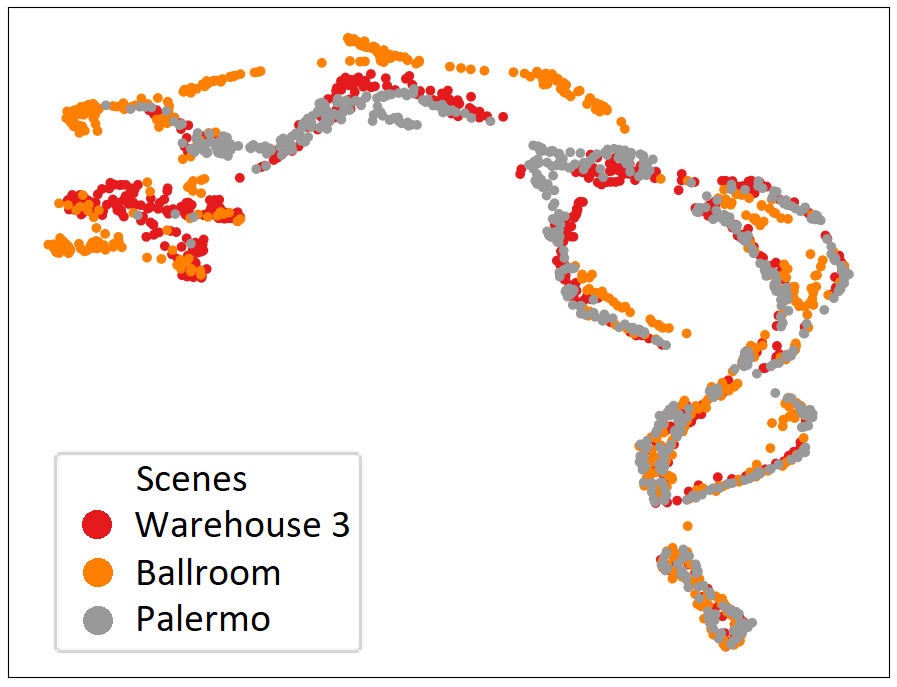}
\caption{UMAP projection of learned visual embeddings from trajectories collected in three held-out environments.}
    \label{fig:umap}
\end{figure}

We visualize embeddings from held-out environment trajectories using UMAP. The three environments share identical trajectory configurations, differing only in visual appearance. Fig.~\ref{fig:umap} shows that observations from visually distinct scenes organize along a coherent latent manifold, with embeddings from different environments interleaved along shared branches rather than forming scene-specific clusters. Notably, trajectory-like structure emerges across environments without explicit temporal or topological regularization. We attribute this to the adaptive temperature mechanism: temporally nearby observations tend to share high geometric similarity in LiDAR space, causing adaptive $\tau$ to soften their mutual repulsion and keeping them naturally close in the latent space. This suggests the representation captures navigation-relevant geometry that generalizes beyond scene appearance.

\subsection{Navigation Performance Under Distribution Shift}
\label{sec:navigation_performance}

We evaluate downstream navigation using a frozen visual encoder for all methods and train a single policy architecture with identical hyperparameters to isolate representation quality. Each method trains with a frozen visual encoder and identical hyperparameters (five seeds) in Photo-studio (Fig.~\ref{fig:data_overview}~(b)), except DR-PR and DR-CSDM which additionally use training and held-out environments respectively. Evaluation uses test environments (Fig.~\ref{fig:data_overview}~(d) \& (e)) over 100 episodes per environment. We report Success Rate (SR) and Success weighted by Path Length (SPL)~\cite{yokoyama2021success}. Results are shown in Table~\ref{tab:all_methods_floor} and Table~\ref{tab:all_methods_in_out}.

\begin{table}
\caption{Navigation performance under increasing appearance and scene shifts. Policies are trained in a single environment and evaluated on unseen test scenes. SR and SPL on the test environments are averaged over 5 seeds and all environments in each class. The best performance are highlighted in bold, and the second best are underlined}
\label{tab:all_methods_floor}
\centering
\scriptsize
\setlength{\tabcolsep}{2pt}
\begin{tabular}{l c cc|cc|cc|cc}
\toprule
\textbf{Method} & \thead{\textbf{Custom}\\\textbf{Dataset}} & \multicolumn{2}{c|}{\thead{\textbf{Train}}} & \multicolumn{2}{c|}{\thead{\textbf{Test}\\\textbf{Simple}}} & \multicolumn{2}{c|}{\thead{\textbf{Test Hard}\\\textbf{Simple Floor}}} & \multicolumn{2}{c}{\thead{\textbf{Test Hard}\\\textbf{Textured Floor}}} \\
\cmidrule(lr){3-4} \cmidrule(lr){5-6} \cmidrule(lr){7-8} \cmidrule(lr){9-10}
 & & SR↑ & SPL↑ & SR↑ & SPL↑ & SR↑ & SPL↑ & SR↑ & SPL↑ \\
\midrule
CL-RN\cite{radford2021learning}  & \texttimes & \underline{92.78} & \underline{91.14} & 38.80 & 33.00 & 33.72 & 28.46 & 10.12 & 8.48 \\
 & & {\tiny ±11.47} & {\tiny ±12.76} & {\tiny ±13.12} & {\tiny ±13.11} & {\tiny ±10.0} & {\tiny ±9.42} & {\tiny ±8.81} & {\tiny ±7.66} \\
CL-VT\cite{radford2021learning}  & \texttimes & 87.16 & 83.25 & \underline{52.00} & \underline{44.66} & 24.07 & 19.41 & 6.9 & 5.59 \\
 & & {\tiny ±5.19} & {\tiny ±6.90} & {\tiny ±8.97} & {\tiny ±4.40} & {\tiny ±9.95} & {\tiny ±8.77} & {\tiny ±7.16} & {\tiny ±6.07} \\
MAE\cite{he2022masked}  & \texttimes & \textbf{93.40} & \textbf{91.48} & 37.20 & 30.22 & 18.22 & 15.28 & 6.82 & 5.93 \\
 & & {\tiny ±2.31} & {\tiny ±2.31} & {\tiny ±15.87} & {\tiny ±12.18} & {\tiny ±9.64} & {\tiny ±8.33} & {\tiny ±7.54} & {\tiny ±6.62} \\
Resnet\cite{he2016deep}  & \texttimes & 87.40 & 82.83 & 44.4 & 34.97 & 25.40 & 18.85 & 5.3 & 3.97 \\
 & & {\tiny ±3.43} & {\tiny ±5.45} & {\tiny ±26.25} & {\tiny ±20.23} & {\tiny ±15.27} & {\tiny ±13.11} & {\tiny ±5.26} & {\tiny ±4.15} \\
DINOv2\cite{oquab2023dinov2}  & \texttimes & 82.00 & 77.16 & 42.80 & 34.15 & 37.35 & 32.33 & 15.2 & 12.77 \\
 & & {\tiny ±3.16} & {\tiny ±4.60} & {\tiny ±14.02} & {\tiny ±14.11} & {\tiny ±5.96} & {\tiny ±6.35} & {\tiny ±8.6} & {\tiny ±7.3} \\

CRL\cite{du2021curious} & \texttimes & 88.4 & 84.47 & 0.0 & 0.0 & 9.45 & 8.34 & 9.77 & 8.59 \\
 & & {\tiny ±5.68} & {\tiny ±5.27} & {\tiny ±0.0} & {\tiny ±0.0} & {\tiny ±8.23} & {\tiny ±7.38} & {\tiny ±6.83} & {\tiny ±6.33} \\

AE\cite{hinton2006reducing} & \checkmark & 85.2 & 81.95 & 0.0 & 0.0 & 0.4 & 0.26 & 0.62 & 0.39 \\
 & & {\tiny ±6.18} & {\tiny ±6.31} & {\tiny ±0.0} & {\tiny ±0.0} & {\tiny ±0.89} & {\tiny ±0.58} & {\tiny ±1.16} & {\tiny ±0.73} \\
MAE\cite{he2022masked} \tablefootnote{Since domain mismatch \ref{sec:csdm} was also used for the MAE fine-tuned, the model overfit to the pretraining environments, resulting in lower performance than the base MAE model even in the training environment.} & \checkmark & 90.8 & 87.18 & 0.4 & 0.11 & 2.27 & 1.91 & 0.85 & 0.71 \\
 & & {\tiny ±4.43} & {\tiny ±3.64} & {\tiny ±0.54} & {\tiny ±0.19} & {\tiny ±4.56} & {\tiny ±3.95} & {\tiny ±1.59} & {\tiny ±1.43} \\
SimCLR\cite{chen2020simple} & \checkmark & 82.2 & 77.86 & 39.4 & 33.36 & 51.07 & 44.76 & \underline{45.7} & \underline{39.39} \\
 & & {\tiny ±9.98} & {\tiny ±10.29} & {\tiny ±14.77} & {\tiny ±10.24} & {\tiny ±8.61} & {\tiny ±7.37} & {\tiny ±12.7} & {\tiny ±11.43} \\

DR-PR & \checkmark & 76.2 & 74.71 & 45.0 & 37.98 & 44.9 & 38.73 & 30.02 & 25.73 \\
& & {\tiny ±7.25} & {\tiny ±6.65} & {\tiny ±19.44} & {\tiny ±15.62} & {\tiny ±11.94} & {\tiny ±9.54} & {\tiny ±10.45} & {\tiny ±9.32} \\

DR-CSDM & \checkmark & 77.4 & 75.66 & 28.2 & 24.46 & \underline{52.52} & \underline{44.86} & 42.22 & 36.47 \\
& & {\tiny ±8.98} & {\tiny ±9.42} & {\tiny ±5.01} & {\tiny ±4.50} & {\tiny ±7.82} & {\tiny ±7.51} & {\tiny ±9.10} & {\tiny ±8.43} \\
\midrule
\textbf{Ours} & \checkmark & 83.8 & 79.62 & \textbf{82.6} & \textbf{78.82} & \textbf{77.7} & \textbf{73.86} & \textbf{49.42} & \textbf{43.36} \\
 & & {\tiny ±7.46} & {\tiny ±7.19} & {\tiny ±5.68} & {\tiny ±5.99} & {\tiny ±4.28} & {\tiny ±4.08} & {\tiny ±11.11} & {\tiny ±9.67} \\
\bottomrule
\end{tabular}
\end{table}
\begin{table}
\caption{Generalization across indoor and outdoor environments. Policies are trained in a single indoor scene and evaluated on unseen environments. Results are averaged over 5 seeds (mean $\pm$ std).}
\label{tab:all_methods_in_out}
\centering
\resizebox{\linewidth}{!}{
\scriptsize
\setlength{\tabcolsep}{0.7pt}
\begin{tabular}{l c cc|cc|cc}
\toprule
\textbf{Method} & \thead{\textbf{Custom}\\\textbf{Dataset}} 
& \multicolumn{2}{c|}{\textbf{Train}} 
& \multicolumn{2}{c|}{\textbf{Indoor}} 
& \multicolumn{2}{c}{\textbf{Outdoor}} \\
\cmidrule(lr){3-4} \cmidrule(lr){5-6} \cmidrule(lr){7-8}
 & & SR↑ & SPL↑ & SR↑ & SPL↑ & SR↑ & SPL↑ \\
\midrule
CL-RN\cite{radford2021learning}  & \texttimes & \underline{92.78} {\tiny ±11.47} & \underline{91.14} {\tiny ±12.76} & 19.15 {\tiny ±10.04} & 16.02 {\tiny ±8.92} & 24.7 {\tiny ±8.84} & 20.92 {\tiny ±8.17}\\
CL-VT\cite{radford2021learning}  & \texttimes & 87.16 {\tiny ±5.19} & 83.25 {\tiny ±6.9} & 17.52 {\tiny ±10.10} & 14.29 {\tiny ±8.50} & 13.45 {\tiny ±7.01} & 10.71 {\tiny ±6.34} \\
MAE\cite{he2022masked}  & \texttimes & \textbf{93.40} {\tiny ±2.31} & \textbf{91.48} {\tiny ±2.31} & 19.22 {\tiny ±9.69} & 16.59 {\tiny ±8.97} & 5.82 {\tiny ±7.49} & 4.62 {\tiny ±5.97} \\
Resnet\cite{he2016deep}  & \texttimes & 87.40 {\tiny ±3.43} & 82.83 {\tiny ±5.45} & 17.97 {\tiny ±11.7} & 13.43 {\tiny ±9.91} & 12.72 {\tiny ±8.82} & 9.4 {\tiny ±7.35} \\
DINOv2\cite{oquab2023dinov2}  & \texttimes & 82.00 {\tiny ±3.16} & 77.16 {\tiny ±4.60} & 38.37 {\tiny ±7.87} & 32.85 {\tiny ±7.68} & 14.17 {\tiny ±6.69} & 12.25 {\tiny ±5.96} \\

CRL\cite{du2021curious}  & \texttimes & 88.4 {\tiny ±5.68} & 84.47 {\tiny ±5.27} & 12.63 {\tiny ±8.39} & 11.23 {\tiny ±7.76} & 8.02 {\tiny ±7.40} & 7.04 {\tiny ±6.74} \\

AE\cite{hinton2006reducing} & \checkmark & 85.2 {\tiny ±6.18} & 81.95 {\tiny ±6.31} & 0.7 {\tiny ±1.49} & 0.49 {\tiny ±1.05} & 0.32 {\tiny ±0.55} & 0.16 {\tiny ±0.26} \\
MAE\cite{he2022masked} & \checkmark & 90.8 {\tiny ±4.43} & 87.18 {\tiny ±3.64} & 1.27 {\tiny ±2.27} & 0.97 {\tiny ±1.8} & 1.85 {\tiny ±3.89} & 1.66 {\tiny ±3.57} \\
SimCLR\cite{chen2020simple} & \checkmark & 82.2  {\tiny ±9.98} & 77.86 {\tiny ±10.29} & \underline{50.14}  {\tiny ±10.17} & \underline{43.56} {\tiny ±9.25} & 47.3 {\tiny ±11.92} & 41.51 {\tiny ±10.43} \\

DR-PR & \checkmark &  76.2 {\tiny ±7.25} & 74.71 {\tiny ±6.65} & 41.50  {\tiny ±9.97} & 35.96 {\tiny ±9.25} & 35.91 {\tiny ±10.57} & 30.81 {\tiny ±9.46} \\

DR-CSDM & \checkmark &  77.4 {\tiny ±8.98} & 75.66 {\tiny ±9.42} & 46.82  {\tiny ±8.79} & 39.86 {\tiny ±7.87} & \underline{49.82} {\tiny ±8.14} & \underline{43.40} {\tiny ±7.99} \\
\midrule
\textbf{Ours} & \checkmark & 83.8 {\tiny ±7.46} & 79.62 {\tiny ±7.19} & \textbf{68.65} {\tiny ±7.18} & \textbf{63.63 }{\tiny ±6.58} & \textbf{60.89} {\tiny ±8.5} & \textbf{55.79} {\tiny ±7.6} \\
\bottomrule
\end{tabular}
}
\end{table}
\begin{table}[t]
\caption{Ablation study of representation learning components.}
\label{tab:ablation}
\centering
\small
\resizebox{0.8\linewidth}{!}{
\setlength{\tabcolsep}{4pt} 
\begin{tabular}{ccccc | cc}
\toprule
\makecell{\textbf{RL} \\ \textbf{DR}} & \makecell{\textbf{Adaptive} \\ $\boldsymbol{\tau}$} & \textbf{CSDM} & \makecell{\textbf{Scene} \\ \textbf{Inv.}} & \makecell{\textbf{Vis.} \\ \textbf{Augs.}} & \textbf{SR $\uparrow$} & \textbf{SPL $\uparrow$} \\
\midrule
\texttimes & \texttimes & \texttimes & \texttimes & \checkmark & 48.38 ±\scriptsize{10.6} & 42.07 ±\scriptsize{9.4} \\
\texttimes & \texttimes & \texttimes &  \checkmark & \checkmark & 43.07 ±\scriptsize{9.18} & 37.77 ±\scriptsize{8.67} \\
\texttimes & \texttimes & \texttimes & \checkmark & \texttimes & 50.20 ±\scriptsize{5.78} & 47.08 ±\scriptsize{5.87} \\

\texttimes & \texttimes & \checkmark & \checkmark & \texttimes & \underline{59.86} ±\scriptsize{7.95} & \underline{55.31} ±\scriptsize{7.47} \\
\texttimes & \checkmark & \checkmark & \checkmark & \texttimes & \textbf{63.56} ±\scriptsize{7.7} & \textbf{58.61} ±\scriptsize{6.87} \\
\midrule
\checkmark & \checkmark & \checkmark & \checkmark & \texttimes & 54.27 ±\scriptsize{6.24} & 49.56 ±\scriptsize{5.81} \\
\checkmark & \checkmark & \texttimes & \checkmark & \texttimes & 51.30 ±\scriptsize{5.22} & 46.04 ±\scriptsize{4.46} \\
\bottomrule
\end{tabular}
}
\end{table}
\subsubsection{In-Distribution Performance}

All methods achieve high SR and SPL (typically $>80\%$ SR) in the training environment (see column ``Train" in Table~\ref{tab:all_methods_floor} or~\ref{tab:all_methods_in_out}), showing the task is solvable with sufficient interaction data. However, strong in-distribution performance does not imply robustness: methods with similar training SR behave very differently under distribution shift.

\subsubsection{Appearance and Scene Shifts}

We first evaluate robustness to controlled visual perturbations (See ``Test Simple" in Table~\ref{tab:all_methods_floor}). Under simple appearance changes, the floor and background are replaced with two uniform colors, introducing a global appearance shift while preserving scene geometry, all the baselines lose 35\%–55\% in SR relative to training. Methods finetuned on the custom dataset (AE, MAE) collapse entirely. In contrast, our method retains 82.6\% SR, maintaining performance close to the training regime.

Under more severe scene changes with fixed floor appearance (Fig. \ref{fig:data_overview} (d) \& (e), but with uniform color as floor as \emph{Test Hard – Simple Floor}), baseline performance further degrades, but simCLR got 51.07\% SR. Our approach achieves 77.7\% SR, substantially outperforming all baselines and reducing the generalization gap by more than half. Notably, neither of the domain-randomization baselines (DR-PR or DR-CSDM) reached the performance of our method, suggesting that increasing scene diversity alone is insufficient to achieve the level of transfer provided by geometry-guided representation learning.

The most challenging setting (\emph{Test Hard – Textured Floor}) introduces the textured floor and additional lighting and reflection changes. All methods degrade, but ours remains markedly more robust (49.4\% SR), achieving more than 3× the performance of most pretrained models (i.e. DINOv2) and clearly outperforming SimCLR (45.7\% SR). Importantly, textured floors are never observed during training, indicating that the learned representation captures navigation-relevant structure rather than superficial visual statistics.

\subsubsection{Indoor-to-Outdoor Generalization}

We next evaluate semantic domain shift by averaging results across indoor and outdoor environments (Table~\ref{tab:all_methods_in_out}). Unlike the controlled appearance perturbations above, indoor-to-outdoor transfer introduces a semantic domain shift. Here, performance differences reflect the extent to which representations encode scene-agnostic navigation structure rather than environment-specific semantics. While pretrained baselines show substantial degradation across domains, our method maintains consistent performance across both indoor and outdoor settings (68.7\% vs. 60.9\% SR), with a significantly smaller generalization gap.

\subsubsection{Ablation Study}

Table~\ref{tab:ablation} evaluates our framework components. We denote: "RL DR" (RL phase domain randomization), "Adaptive $\tau$" (geometry-adaptive temperature, Sec.~\ref{sec:adaptive_tau}), "CSDM" (environment mismatch between pretraining and policy learning, Sec.~\ref{sec:csdm}), "Scene Inv." (scene-invariant positive pairs), and "Vis. Augs." (standard SimCLR augmentations). The SimCLR baseline achieves 48.4\% SR. Scene invariance alone improves over visual augmentations in isolation, but combining them degrades performance, suggesting standard SimCLR augmentations may be misaligned with geometry-oriented contrastive learning for navigation. CSDM consistently yields substantial gains across configurations. Comparing rows 3–4 and rows 6–7, CSDM improves performance by +9.7\% SR and +3.0\% SR, respectively, demonstrating that viewpoint diversity via cross-stage domain mismatch is robust to other design choices. This consistency underscores the fundamental importance of domain mismatch in learning robust representations. Adaptive temperature further improves to 63.6\% SR and 58.6\% SPL. Overall, robustness is driven primarily by domain mismatch and adaptive weighting, rather than conventional augmentations.

\subsubsection{Discussion: Cross-Stage Domain Mismatch vs. Domain Randomization}

We investigated domain randomization as an alternative approach to robustness, testing it on both our final model and the SimCLR baseline (Table~\ref{tab:all_methods_in_out} and ~\ref{tab:ablation}). Surprisingly, DR does not improve performance: on our model, DR reduces performance from 63.6\% to 54.27\% SR; on SimCLR, DR-PR underperforms the baseline (48.38\% → 47.37\% on average of all unseen scenes).
This reveals a limitation: domain randomization's effectiveness depends on whether the randomization covers the actual domain shift. In our case, training environments contained uniform-colored floors, so the randomization lacked textured floor configurations. This mismatch caused overfitting to constant colors, degrading performance on textured floors—a shift larger than the randomization could encompass.
In contrast, CSDM works through an implicit mechanism that does not require domain knowledge. The frozen encoder forces the policy to rely on geometry-consistent features rather than appearance-specific shortcuts. CSDM thus acts as an implicit regularizer promoting usefulness-based feature selection, explaining why it consistently improves performance (rows 3–4, 6–7, Table~\ref{tab:ablation}) regardless of domain randomization.

\subsection{Sim2Real Transfer}

To evaluate sim-to-real transfer, the trained policy was deployed on a Limo Pro mobile robot equipped with an RGB camera and onboard computing. Due to computational limitations, policy inference was performed on an external laptop, and the generated actions were transmitted to the robot in real time.

\begin{figure}[]
    \centering
\includegraphics[width=0.4\textwidth]{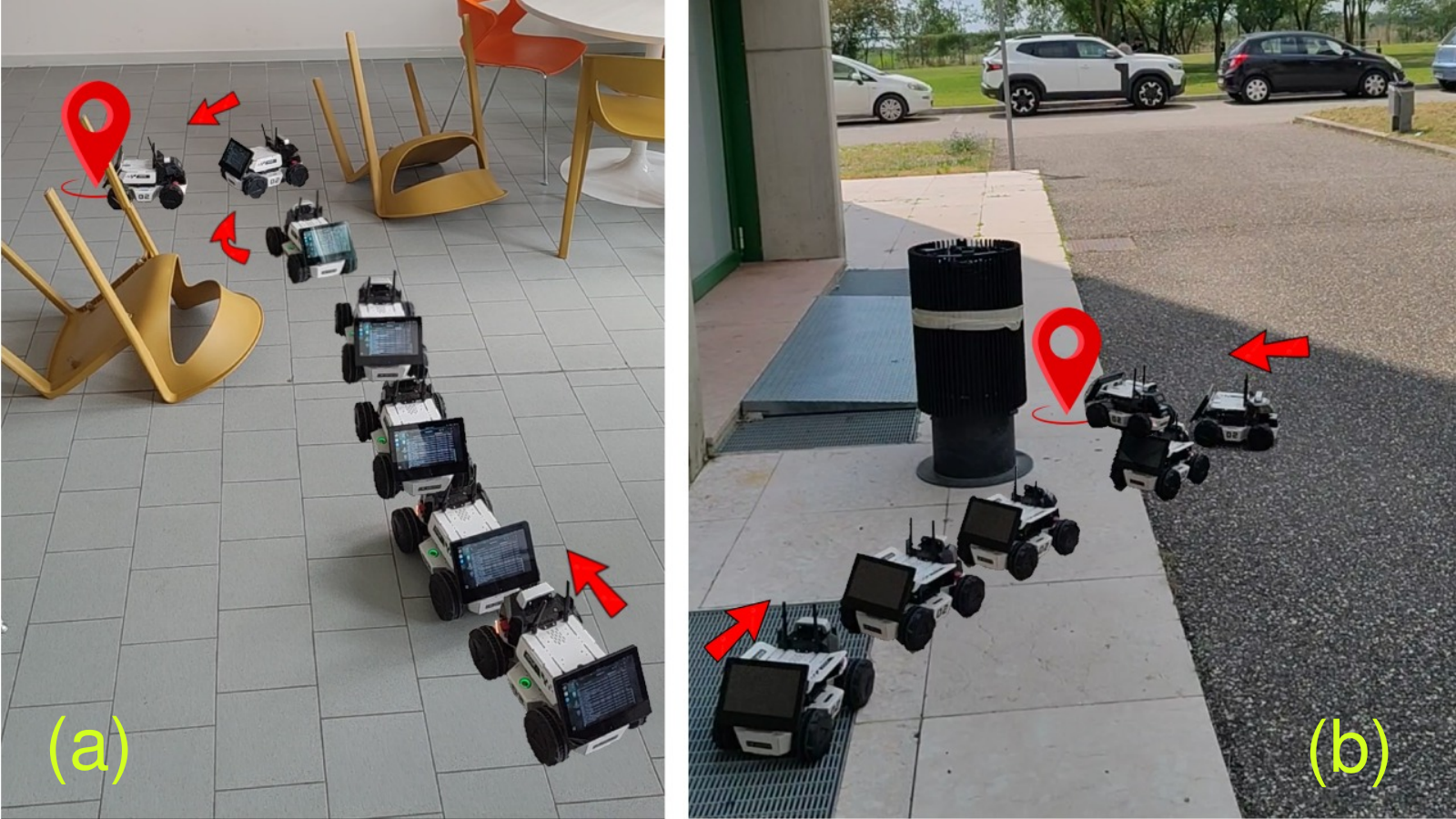}
\caption{Composite image from the real-world experiments showing the robot navigation. (a) outdoor, (b) indoor}
    \label{fig:real_world}
\end{figure}

The pretrained policy demonstrated zero-shot transfer capability and was directly deployed on the physical robot without any real-world fine-tuning. We evaluated the final model in ten navigation trials across indoor and outdoor environments, as shown in Fig.~\ref{fig:real_world}. The policy achieved success rates of 60\% indoors and 40\% outdoors. Given the limited number of trials, these results should be interpreted as a proof-of-concept demonstration rather than a statistically exhaustive evaluation. Failures were primarily caused by collisions with previously unseen obstacles, such as a small square curb in the outdoor environment. Additionally, occasional backward motion was observed near the goal. We attribute this behavior to the reward design in Eq.~\ref{eq:reward}, where equal penalties for backward and angular motion can make reverse correction locally preferable to reorientation. Overall, the results demonstrate the feasibility of zero-shot sim-to-real transfer while highlighting the importance of obstacle diversity during training and reward coefficient tuning for robust real-world deployment.

\section{Conclusion}
We presented a sensor-guided contrastive representation learning framework for robust scene transfer in vision-based PointGoal navigation. By leveraging privileged geometric sensing only during training to adaptively modulate the contrastive objective, our method learns navigation-relevant visual representations invariant to scene appearance. Decoupling representation learning from policy optimization further promotes reliance on task-relevant features.

Extensive experiments demonstrate substantial policy-level generalization across unseen indoor/outdoor environments under severe appearance shifts. At deployment, the agent uses only monocular RGB observations and standard inputs (goal position, proprioceptive signals), without privileged sensors. These results validate privileged-guided contrastive learning as an effective mechanism for improving robustness in vision-based navigation.

Future work will explore: (1)~reducing reliance on explicit task inputs like goal position, (2)~interaction between geometry-aware objectives and image augmentation strategies, and (3)~which privileged sensors most effectively guide representation learning.

\bibliographystyle{IEEEtran}  
\bibliography{biblio}

\end{document}